\documentclass[twoside]{article}
\usepackage[accepted]{aistats2023}
\usepackage{styles/user_commands}
\usepackage{listings}
\usepackage[linesnumbered,lined,commentsnumbered, ruled, vlined ]{algorithm2e}  
\usepackage{tikz,pgfplots}
\pgfplotsset{compat=1.6}
\usetikzlibrary{shapes,positioning}
\usepackage{standalone}
\graphicspath{{pictures/}{../pictures/}}
\usepackage{subcaption}
\usepackage{tabularx}
\usepackage{booktabs}

\usepackage{subfiles}
\usepackage{algorithmic}
\usepackage[cal=boondox]{mathalfa}

\usepackage{amsmath,enumerate,amsbsy,amsfonts,amssymb,amscd,graphicx,mathabx}
\usepackage[round,authoryear]{natbib}
\usepackage{authblk}
\usepackage[utf8]{inputenc}
\usepackage{graphbox}
\usepackage{multirow,graphicx, xcolor}
\usepackage{booktabs}
\usepackage{tabularx}

\usepackage[T1]{fontenc}    % use 8-bit T1 fonts
\usepackage{hyperref}       % hyperlinks
\usepackage{url}            % simple URL typesetting
\usepackage{booktabs}       % professional-quality tables
\usepackage{amsfonts}       % blackboard math symbols
\usepackage{nicefrac}       % compact symbols for 1/2, etc.
\usepackage{microtype}      % microtypography
\usepackage{xcolor}         % colors
\usepackage{dsfont}            % indicator
\def\T{{ \mathrm{\scriptscriptstyle T} }}

\begin{document}
\runningauthor{Yirui Liu, Xinghao Qiao, Liying Wang, Jessica Lam}

\runningtitle{Edge Enhanced Graph Neural Network}

\twocolumn[

\aistatstitle{EEGNN: Edge Enhanced Graph Neural Network with a Bayesian Nonparametric Graph Model}

\aistatsauthor{ Yirui Liu$^{1, 2}$ \And Xinghao Qiao$^{1}$ \And  Liying Wang$^{3}$  \And Jessica Lam$^{4}$ }

\aistatsaddress{$^{1}$London School of Economics and Political Science\\
$^{3}$Bayes Business School, City, University of London \And 
$^{2}$J.P. Morgan\\ 
$^{4}$University of Oxford
} ]

\begin{abstract}
Training deep graph neural networks (GNNs) poses a challenging task, as the performance of GNNs may suffer from the number of hidden message-passing layers. 
The literature has focused on the proposals of {over-smoothing} and {under-reaching} to explain the performance deterioration of deep GNNs.
In this paper, we propose a new explanation for such deteriorated performance phenomenon,  {mis-simplification}, that is, mistakenly simplifying graphs by preventing self-loops and forcing edges to be unweighted. We show that such simplifying can reduce the potential of message-passing layers to capture the structural information of graphs.
In view of this, we propose a new framework, edge enhanced graph neural network (EEGNN). EEGNN uses the structural information extracted from the proposed Dirichlet mixture Poisson graph model (DMPGM), a Bayesian nonparametric model for graphs, to improve the performance of various deep message-passing GNNs. We propose a Markov chain Monte Carlo inference framework for DMPGM. Experiments over different datasets show that our method achieves considerable performance increase compared to baselines. 
%An empirical study with financial data demonstrates that EEGNN can improve the performance of GNN-based investment strategies.

\end{abstract}

\section{INTRODUCTION}
\label{sec:intro}
Graph neural networks (GNNs) \cite[]{zhou2020, wu2020} are important tools for analyzing graph data, such as social network \cite[]{you2020}, transportation network \cite[]{chen2021a}, molecular graph \cite[]{huang2020}, biological network \cite[]{zhang2021}, financial transaction network \cite[]{wang2021}, academic citation graph \cite[]{xu2021a}, and knowledge graph \cite[]{ji2021}. 
GNNs have become popular with their state-of-the-art performance by applying deep learning methodologies to graphs. Among them, message passing neural networks (MPNN) \cite[]{gilmer2017} uses message-passing layers to compute node embeddings. Examples of MPNNs include graph convolutional neural networks (GCN) \cite[]{kipf2017a}, GraphSAGE \cite[]{hamilton2017}, graph attention networks (GAT) \cite[]{velickovic2018}, and gated graph neural networks (GGNN) \cite[]{li2016}. Similar to standard multi-layer perceptron (MLP) in deep learning, the message passing layer in a GNN framework aggregates information from the local neighbors of each node, and then transforms the information via an activation function into the embedding \cite[]{hamilton2020}. A node embedding can aggregate information over $N$ hop neighbors, in the form of $N$ hidden message-passing layers, thus incorporating further reaches of the graph.

Although deeper layers in non-graph neural networks often achieve better performance \cite[]{krizhevsky2012, he2016}, GNNs typically perform best with only 2 to 4 hop neighbors, that is, 2 to 4 hidden layers. In contrast, using a larger number of layers, termed as deep stacking, may lead to a substantial drop in the performance for GNNs \cite[]{Klicpera2019, rong2019, li2020, chen2020a}. One explanation for this phenomenon is the \textit{over-smoothing}. By applying graph convolution repeatedly over many hidden layers, the representation of the nodes will be indistinguishable. As a result, the \textit{over-smoothing} can jeopardize the  performance of deep GNNs. Another explanation is the \textit{under-reaching}. When GNNs aggregate messages over long paths, the information propagation across distant nodes in the graph becomes difficult because it is susceptible to bottlenecks \cite[]{alon2020}. This causes GNNs to perform poorly in predicting tasks that require remote interaction \cite[]{singh2021, hwang2021}.
						
Many efforts have been devoted to addressing these limitations. To handle the {over-smoothing}, DropEdge \cite[]{rong2019} and DropNode \cite[]{huang2021} were proposed to randomly remove a certain number of edges or nodes from the input graph at each training epoch. These methods are likened to Dropout \cite[]{srivastava2014}, which randomly drops hidden neurons in neural networks to prevent overfitting. On the contrary, to address the {under-reaching}, virtual edges \cite[]{gilmer2017}, super nodes \cite[]{scarselli2009, hwang2021}, or short-cut edges \cite[]{allamanis2018} can be added to the original graph. However, none of the aforementioned methods consider adding or removing based on the structural information of the graph. Instead, the pattern of deciding which nodes or edges to be added or removed comes from an arbitrarily random selection. Although dropout has been effective in non-graph neural networks, its random removal and addition of nodes can disturb the graph structure, thus compromising the performance of GNNs that relies on the structure to propagate information.
%that the observed neighborhood has limited representation power and cannot learn basic topological properties of the graph \cite[]{vignac2020}. 

\begin{figure}
  \centering
  \begin{subfigure}{0.48\textwidth}
         \centering
\includegraphics[width=0.75\textwidth]{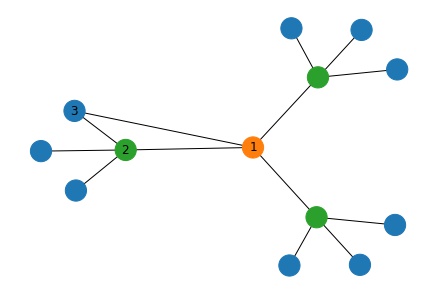}
         \caption{The observed simple graph}
         \label{fig:subfig_0}
     \end{subfigure}
  \begin{subfigure}{0.48\textwidth}
  \centering
  \includegraphics[width=0.75\textwidth]{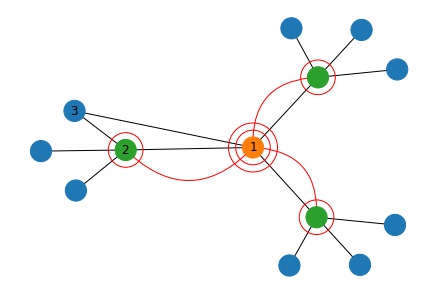}
         \caption{The corresponding virtual multigraph}
         \label{fig:subfig_1}
     \end{subfigure}
  \caption{The observed simple graph versus the virtual multigraph. The red edges are virtual edges in the virtual multigraph. In particular, the red circles are virtual self-loops.}
   \label{fig:fig_1}
\end{figure}

Different from the \textit{over-smoothing} and \textit{under-reaching}, we propose a new explanation for performance deterioration of deeper GNNs from the perspective of misusing edge structural information, \textit{mis-simplification}, explained as follows. Most observed graphs are recorded as simple graphs, where self-loops are not allowed, and all edges are unweighted and undirected \cite[]{shafie2015}. In a natural way, GNNs are designed for learning such simple graphs that can be constructed by collapsing multiple edges into a single edge as well as removing self-loops. This approach, however, discards the information inherent in the original network. Take one example, for a source node connected to many neighboring target nodes (see node $1$ in Figure~\ref{fig:subfig_0}), its self-loop has an equal weight to neighboring non-loop edge, which may under-weigh the importance of this node. Take another example, no matter how similar the two nodes are (see nodes $1$ and $2$ in Figure~\ref{fig:subfig_0}), only one edge is allowed to connect the pair of nodes, and as a result, the information passing between both nodes is restricted. Furthermore, edge $(1,2)$ should play a more important role in message passing than edge $(1,3)$, because node $2$ is a key node with 3 sub-nodes in total,  while node $3$ is just a sub-node of node $2$. However, typical GNNs treat these two edges indifferently as they are equally weighted in the simple graph.
Therefore, such \textit{mis-simplification} can reduce the potential of message-passing layers to capture structural information in GNNs.

To solve this issue, we propose an edge-enhanced graph neural network (EEGNN), which incorporates edge structural information in the message-passing layer. First, we assume that there is an underlying \textit{virtual multigraph}, allowing for self-loops and for multiple edges between pairs of nodes, and the observed graph model can be viewed as a transformation of the virtual multigraph. As illustrated in Figure \ref{fig:fig_1}, the above Figure~\ref{fig:subfig_0}is the original observed simple graph, while the below Figure~\ref{fig:subfig_1} is the corresponding virtual graph.
Second, to build the virtual multigraph that can capture the edge structural information, we propose the Dirichlet mixture Poisson graph model, a Bayesian nonparametric model. Following \cite{caron2017}, the interactions between nodes are modelled by assigning a \textit{sociability} parameter to each node. Then, the counts of edges are generated from  a Poisson distribution, where the Poisson rate is the product of sociability parameters of the nodes in two ends. Finally, in the framework of EEGNN, we can then replace the observed graph in a GNN with the virtual multigraph. 
%For example, an edge enhanced GCN uses the virtual multigraph instead of the observed graph for the convolution. 
In this architecture, message-passing layers can then assign weights proportionally to the importance of the edges, thus passing the information from nodes to nodes in a more reasonable manner.

%In this paper, we provide the following key contributions:
The main contribution of our paper is fourfold.
\begin{itemize}
\item We outline a new explanation for the poor performance of deep GNNs;
\item We propose a new way to enhance existing GNN methods by utilizing the structural information of graphs;
\item We propose a Bayesian nonparametric graph model and its Monte Carlo Markov chain (MCMC) inference procedure;
\item We demonstrate the superior sample performance of our proposal over existing methods through the experiments on six real datasets and a financial application.
%conduct an empirical study to demonstrate that our proposed method can enhance existing methods performace on real datasets.
\end{itemize}

\section{Preliminaries}
\subsection{GNN and Message Passing Layer}
We begin by introducing some notation. Let $G = (V, E)$ be a
graph with node set $V = \{v_1,\dots, v_{|V|}\}$ and edge set $E=\{e_1,\dots, e_{|E|}\}$, where $|V|$ and $|E|$ denote the number of nodes and edges in $G$, respectively. The adjacency
matrix $A \in \R^{|V|\times |V|}$ is defined as $A_{ij} = 1$ if $(v_i,v_j) \in E$
and $0$ otherwise. The corresponding degree matrix $D \in  \R^{|V| \times |V|}$ is defined as
$D = \text{diag}(D_1,\dots,D_{|V|})$, where $D_i = \sum_{j=1}^{|V|} A_{ij}$. 
We denote the data matrix by $X \in R^{m\times |V|},$ whose $j$-th column corresponds to a $m$-dimensional feature vector of node $j$. 
%denotes the feature matrix with the features for each node corresponds to a column in $X$, where $m$ is the dimension of feature space. 

GNN is a neural network model to process graphs for node classification, edge prediction and  graph classification \cite[]{gori2005, zhou2020, wang2021}. Within various GNNs, information is exchanged between nodes and is updated by neural networks via message passing layers \cite[]{gilmer2017}. Specifically, the initial representation, $h_i^0$ for node $i$, is generated by a function of this node's features. Then, the message passing layers update the representation based on this node's neighbors. The message passing contains two steps: the aggregation step and the update step.  Denote the representation for node $i$ in layer $l$ by $h_i^{l}.$ A message passing layer in GNN can be expressed as
\begin{equation}
    h_i^{l+1} = \text{UPDATE}(h_i^{l}, \text{AGGREGATE}(h_j^{l} \mid j \in \cN_i)),
\end{equation}
where $\text{AGGREGATE}(\cdot)$ denotes a permutation-invariant function, such as the sum, mean, and maximum, to send information from one node to another through edges, and $\text{UPDATE}(\cdot)$ denotes linear or nonlinear differentiable functions such as MLP, $\cN_i$ denotes the neighborhood of node $i$, that is, the set of nodes directly connected to node $i$. For example, the vanilla GCN uses $h_i^{l+1} = \sigma(\sum_{j=1}^{|V|}\tilde{P}_{ij} h_j^{l+1} W^l  \mid j \in \cN_i \cup \{i\})$, or in matrix form, $H^{l+1} = \sigma(\tilde{P}H^l W^l)$ \cite[]{kipf2017a},  APPNP uses $H^{l+1} = (1-\alpha)\tilde{P}H^l + \alpha H^0$ \cite[]{Klicpera2019}, and GCNII uses $H^{l+1} = \sigma\big((1-\alpha)\tilde{P}H^l + \alpha H^0)((1-\beta)I_n + \beta W^l)\big)$ \cite[]{chen2020a}, where $\tilde{P} = (D+I)^{-\frac{1}{2}}(A+I)(D+I)^{-\frac{1}{2}}$, $H^l$ is the representation for all nodes in layer $l$, $\sigma$ and $W^l$ are respectively the activation function and the corresponding weight in a neural network layer, and $\alpha$ and $\beta$ are hyperparameters. The formulas above show that the GNN treats each edge with equal weight and hence leads to \textit{mis-simplification}. In order to solve this issue, we adopt a Bayesian nonparametric sparse graph model to generate the virtual edges and virtual multigraph.

\begin{figure}[h]
  \centering
  \begin{subfigure}{0.495\textwidth}
         \centering
\includegraphics[align=c, width=0.65\textwidth]{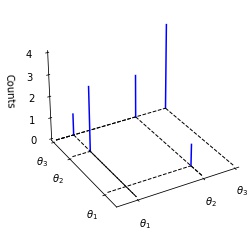}
      \centering
\includegraphics[align=c, width=0.30\textwidth]{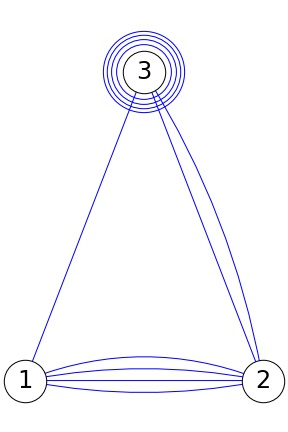}
\caption{Graph model in \cite{caron2017}}
\label{fig:subfig_caron}
\end{subfigure}
\begin{subfigure}{0.495\textwidth}
         \centering
\includegraphics[align=c, width=0.65\textwidth]{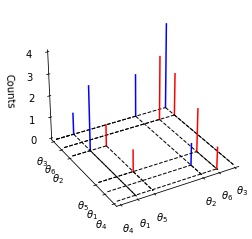}
\includegraphics[align=c, width=0.33\textwidth]{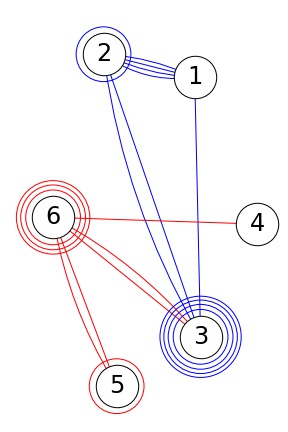}
\caption{Dirichlet mixture Poisson graph model (DMPGM)}
\label{fig:subfig_dmpgm}
     \end{subfigure}
\caption{Bayesian nonparametric graph model. Figure~\ref{fig:subfig_caron} illustrates the model in \cite{caron2017}, while Figure~\ref{fig:subfig_dmpgm} illustrates the proposed graph model in this paper. The left sub-figures show the proxy for nodes,  $\theta_i$s, and the number of edges among them. The right sub-figures display the corresponding multigraph. In Figure~\ref{fig:subfig_dmpgm}, red and blue are used to indicate two clusters in the edges. The circles around nodes denote self-loops. Finally, the number of circles or links denotes the multiplicity.}
\end{figure}

\subsection{Bayesian Nonparametric Sparse Graph Model}
In contrast with other graph models that are based on node feature embeddings, \cite{caron2017} represent the observed graph $G$ as a point process on $\R^2$, $G=\sum_{i,j} z_{i,j}\delta_{(\theta_i, \theta_j)}$, where Dirac function $\delta_{(\theta_i, \theta_j)}$ is equal to $1$ at $(\theta_i, \theta_j)$ and equal to $0$ elsewhere, $z_{i,j}$ is the multiplicity for edge $(i,j)$, and $\theta_i$ is a proxy for node $i$ on the real axis, as illustrated in Figure~\ref{fig:subfig_caron}. Note that the same definition for node $i$ is also applied to node $j$, but we omit the explanation for node $j$ to avoid redundancy. This representation specifies the source and target nodes for each edge. To model the possibility for two nodes constructing an edge, a sociability parameter $w_i>0$ is assigned to node $i$ for each $i=1,\dots, |V|$. Following \cite{aldous1997}, the graph model can be factorized as $p(A_{ij}=1) = 1 - \exp(-2w_iw_j)$ for $i\neq j$ and $p(A_{ii}=1) = 1 - \exp(-w_i^2)$ otherwise. This is equivalent to modelling an unobserved integer-valued multigraph as $z_{ij} \sim \text{Poisson}(w_iw_j)$ and setting $A_{ij} = \mathds{1}_{z_{ij} + z_{ji}>0}.$ To model the sparsity property in real graphs, that is, $|E| = o\big(|V|^2\big)$, the sociability is generated from a completely random measure with infinite activity \cite[]{caron2017}, such as gamma process,  stable process and inverse Gaussian process \cite[]{ghosal2017}. See also Appendix~\ref{app:cmr} for a short review of completely random measures. This model allows for self-loop and multi-edges, and thus can be used to build a virtual multigraph.
However, as the Poisson intensity is factorized as the outer product of a vector with itself, only one feature for each node is considered, which restricts the capability of this model in confronting real data. To address this issue, we propose a novel model in Section~\ref{subsec:dmpgm} below.

\section{METHODOLOGY}
\label{sec:methodology}
\subsection{Dirichlet Mixture Poisson Graph Model}
\label{subsec:dmpgm}
To adopt the latent community information among nodes in the graph, mixed-membership stochastic block model \cite[]{airoldi2008} associates each node with latent cluster distributions.
In an analogy, we add cluster-membership features to each pair of edges instead of nodes. Specifically, we extend the graph model in \cite{caron2017} by proposing the following Dirichlet mixture Poisson graph model (DMPGM),
\begin{equation}
\label{eq:construction}
\begin{gathered}
    \pi=(\pi_1, \pi_2, \cdots) \sim \text{GEM}(\alpha), \\
    W_0=\sum_{i=1}^{\infty} w_{0,i} \delta_{\theta_i} \sim \text{CRM}(\kappa,v), \\%~~ 
    W_k =\sum_{i=1}^{\infty} w_{k,i} \delta_{\theta_i} \sim \Gamma\text{P}(W_0), \\
    z_{ij} \sim \text{Poisson}(\sum_{k=1}^{\infty}\pi_k w_{k,i} \times w_{k,j}), \\%~~ 
    A_{ij} = \text{min}(z_{ij} + z_{ji}, 1) \mathds{1}_{i \neq j},
\end{gathered}
\end{equation}
for $i,j, k \in \N^+$, where $\text{GEM}(\alpha)$ is the distribution for atom sizes of a Dirichlet process $\text{DP}(\alpha)$ and each atom corresponds to a distinct cluster. Moreover, $\text{CRM}(\kappa,v)$ denotes a completely random measure with $v^c(dw, d\theta)=\kappa(d\theta)v(dw)$ as its Levy measure, and $\Gamma\text{P}(H)$ denotes gamma process with the base measure $H$. See Appendix~\ref{app:cmr} for details of these stochastic processes. We summarize the probabilistic generative steps as follows. 
First, the cluster distribution $\pi$ is assigned with a prior $\text{GEM}(\alpha)$, which allows for infinitely many clusters. 
%Conditional on this proportion, for each node pair $(i,j)$, a cluster indicator $c_{ij} \in \N$ is generated from the multinomial distribution. 
Second, we use a hierarchical structure to generate values for the node sociability parameter in each cluster. $W_0$, sampled from a completely random measure, is used as the base measure in $\Gamma\text{P}(W_0)$ for $W_k$ such that $w_{k,i}$ belongs to gamma distribution parameterized by $w_{0,i}$, $w_{k,i} \sim \text{Gamma}(w_{0,i}).$ This hierarchical setting is designed to ensure the components in $W_k$ share atom locations \cite[]{teh2006, liu2020}.
Finally, following \cite{caron2017}, an undirected multigraph $\sum_{i,j} z_{ij}\delta{(\theta_i, \theta_j)}$ is generated from a Poisson process, where $z_{ij}$ is the Poisson-distributed multiplicity for edge $(i,j)$. By aggregating multiple edges to a single edge for each pair of nodes and  removing self-loops, a simple graph is transformed from the multigraph. The corresponding adjacent matrix $A=(A_{ij})$ to the observable simple graph can then be generated. 
An example of DMPGM is illustrated in Figure~\ref{fig:subfig_dmpgm}.

DMPGM can be equivalently expressed under a mixture model framework. Specifically, a set of edges in each cluster is sampled from $\text{Poisson}(\pi_k \widebar{w}_k^2)$, where $\widebar{w}_k = \sum_{i=1}^{\infty}w_{k,i}$. As a consequence, this is equivalent to sampling the total number of edges $n$ from $\text{Poisson}(\lambda)$ with $\lambda = \sum_{k=1}^{\infty}\pi_k \widebar{w}_k^2, $ and then assigning each edge a cluster membership from $\text{Categorical}\big(\frac{\pi_1 \widebar{w}_1^2}{\lambda}, \frac{\pi_2 \widebar{w}_2^2}{\lambda}, \dots\big)$. Following the same methodology, for each edge, a pair of nodes is then sampled from $\text{Categorical}(\frac{w_{k,1}}{\widebar{w}_k}, \frac{w_{k,2}}{\widebar{w}_k}, \dots)$ in the cluster $k$. Hence, a relationship between edges and nodes is constructed. We summarise this equivalent expression for DMPGM as follows,
\begin{equation}
\label{eq:equivalent_construction}
\begin{gathered}
n \sim \text{Poisson}(\sum_{k=1}^{\infty}\pi_k \widebar{w}_k^2), \\
    k \sim \text{Categorical}\big(\frac{\pi_1 \widebar{w}_1^2}{\lambda}, \frac{\pi_2 \widebar{w}_2^2}{\lambda}, \dots\big),\\
    i, j \sim \text{Categorical}\big(\frac{w_{k,1}}{\widebar{w}_k}, \frac{w_{k,2}}{\widebar{w}_k}, \dots\big),
\end{gathered}
\end{equation}
where other structures in equation~\eqref{eq:construction} remain the same.
In Appendix~\ref{app:sec_proof}, we show that DMPGM enjoys similar properties as the model in \cite{caron2017} in the following theorem. % with its proof provided in .
\begin{theorem}
\label{theo:theorem_1}
The graph constructed by DMPGM is sparse if CRM in (\ref{eq:construction}) has infinite activity.
\end{theorem}
For example, using the gamma process as the completely random measure leads to a sparse graph in the DMPGM, which makes it more effective for modeling real-life data.

It is worth noting that DMPGM extends the model in \cite{caron2017} by assuming that edges can belong to different clusters. As a result, DMPGM is more flexible and applicable in modelling real data. We also note that DMPGM is distinct from the overlapping communities graph model \cite[]{todeschini2020} and graph Poisson factorization \cite[]{zhou2015}, because we assign a Dirichlet prior for the clustering distribution, and hence can allow a nonparametric estimation of the number of edge clusters. Moreover, \cite{williamson2016} uses the hierarchical Dirichlet process (HDP) \cite[]{teh2006} to construct the graph. However, as HDP only models the node distribution within a cluster, the number of edges is ignored. As a result, this model cannot be used for EEGNN framework. Finally, 
a generative model that shares some fundamental similarities with DMPGM is proposed by \cite{ricci2022thinned}. However, this concurrent work does not investigate the use of a Bayesian nonparametric graph model to improve GNNs.

\subsection{MCMC Inference Framework}
\label{subsec:inference}
We propose a detailed MCMC framework to infer the posteriors for DMPGM in a nonparametric way. Following \cite{caron2017} and \cite{liu2020}, the posterior distribution for $W_k=\sum_{i=1}^{\infty} w_{k,i} \delta_{\theta_i}$, $k\geq0$, are restricted to the weights $\{w_{k,i}\}$ because the locations $\{\theta_i\}$ of both observed and unobserved nodes are not likelihood identifiable, thus being ignored. Moreover, given the observed nodes set $V$, the weights for each $W_k$ are truncated to a $(|V|+1)$-dimensional vector, $\bw_k = (w_{k, 0}, w_{k, 1},\dots, w_{k, |V|})^\T$, where $w_{k, i}$ corresponds to the weight on an observed node $i$ for $1\leq i \leq |V|$, and $w_{k, 0}$ is the sum of weights for all unobserved nodes.  Similarly, the posterior distribution for $\pi$ is truncated to a $(K+1)$-dimensional vector, $\bpi=(\pi_0, \pi_1, \dots, \pi_K)^\T$, where $K$ is the truncated number of clusters and is inferred adaptively in Step~4 below, and $\pi_0$ corresponds to the cluster without any observation. Consequently, given the truncation levels $|V|$ and $K$ for $W_0$ and $\pi$, respectively, DMPGM contains the following parameters to infer: $\bpi$, $\{\bw_k\}_{k \geq 0} $, $\bz=\{z_{ij}\}_{A_{ij} =1}$ and cluster membership $\bc=\{c_{ijl}\}_{A_{ij} =1, 1 \leq l \leq z_{ij}}$.

We next propose a MCMC inference framework that can infer the number of edge clusters in a Bayesian nonparametric manner in the following steps.
\begin{description}
\item[Step 1] Update $w_{0,1}, \dots, w_{0, |V|} | \widebar{w}_{0}, \bz, \bc$ using Hamiltonian Monte Carlo \cite[]{kroese2011}, where $\widebar{w}_{0} = \sum_{i=0}^{|V|}w_{0,i}$ with the log-posterior and its gradient provided in Appendix~\ref{app:step_1}.

\item[Step 2] Update $w_{k,1},\dots, w_{k,|V|} | \widebar w_{k}, w_0, \bz, \bc$ for $k=1,\dots, K$ given the conjugacy, where $\widebar{w}_{k} = \sum_{i=0}^{|V|} w_{k,i}.$ We sample $\widetilde w_{k, i} \sim  
\text{Dirichlet}(\nu_0, \nu_1, \dots, \nu_{|V|})$, where $\nu_i = w_{0, i} + \sum_{j=1}^{|V|} n_{k,i}$,  $n_{k,i}=\sum_{j=1}^{|V|} \sum_{l=1}^{z_{ij}} \{\mathds{1}_{c_{ijl}=k} +  \mathds{1}_{c_{jil}=k}\}$, and then compute $w_{k, i} = \widebar{w}_{k} \widetilde{w}_{k, i}.$
\item[Step 3] Update $\pi_0, \pi_1,\dots, \pi_K | \bz, \bc$ using the conjugacy. Analogous to Step~2, we sample ${\pi_{k}} \sim  
\text{Dirichlet}(n_0, n_1, \dots, n_k),$ where $n_k = \sum_{i=1}^{|V|} \sum_{j=1}^{|V|}\sum_{l=1}^{z_{ij}} \mathds{1}_{c_{ijl}=k}$  for $k>0$ and $n_0 = \alpha.$

\item[Step 4] Update the latent edge cluster membership  $c_{ijl} | \{\bw_k\}_{k \geq 0}, \bpi$ for each pair $(i, j)$ such that $A_{ij}=1$ and for $l=1,\dots, z_{ij}$. For each edge we sample from the multinomial distribution $p(c_{ijl}=k) ~\propto~ \pi_k w_{k, i} w_{k, j}$ for $k=0,1\dots, K$.  In this step, if $k=0$ is sampled, we add a new cluster \cite[]{teh2006, bryant2012, liu2020}, and increase the truncated number of clusters from $K$ to $K+1$.

\item[Step 5] Update the unobserved $z_{ij} \mid  \bpi, \bw_k \sim \text{Truncated--Poisson}(\sum_{k=0}^{K}\pi_k w_{k,i}w_{k,j})$ for each pair $(i, j)$ such that $A_{ij}=1$, where truncated Poisson is a conditional probability distribution of a Poisson-distributed random variable with strictly positive counts \cite[]{cohen1960}.

\item[Step 6] Update the $\widebar{w}_{k}$ and $\widebar{w}_{0}$ using Metropolis–Hastings \cite[]{kroese2011} algorithm based on  the log-posterior provided in Appendix~\ref{add:step_6}.

\end{description}

We iterate over {Steps 1--6} until convergence. For the MCMC algorithm, the global variables are updated in linear time, and the Monte Carlo step iteratively samples 
from $K$ clusters. Therefore, the computational complexity is dominated by $O\big(K\max\{|V|, |E|\}\big)$.

\subsection{Edge Enhanced Message Passing}
\label{subsec:eemm}
In conventional message passing layers built from a simple graph, information for node $i$ is obtained from edges connected to its neighboring nodes in $\cN_i$ and from its self-loop. In these layers, each edge $(i, j)$ for $j \in \cN_i \cup \{i\}$ has equal weight, resulting in \textit{mis-simplification} of the more complex structural information for the GNN, as described in Section~\ref{sec:intro}. To overcome this \textit{mis-simplification}, we sample  artificial edges given the estimated DMPGM, from which we construct a virtual multigraph
\begin{equation}
    G^* = (V, E, r), ~~r\big((i, j)\big) = z_{ij},
\end{equation}
where the multiplicity-map $r: E \to \N^+$ assigns to each edge an integer to represent its multiplicity, and $z_{ij}$ in DMPGM is defined in (\ref{eq:construction}) and is inferred from Step~5 in Section~\ref{subsec:inference}. In this way, we can extract the edge structural information, via the inferred multiplicity for each edge, using the DMPGM model to build the virtual multigraph. For example, as illustrated in Figure~\ref{fig:fig_1}, two artificial self-loops are added to nodes $1$,  one artificial self-loop is added to nodes $2$, and the edge $(1,2)$ is assigned with multiplicity $2$, where the multiplicity is determined by $z_{11}=2$, $z_{22}=1$ and $z_{12}+z_{21}=2$, respectively.  
We then replace the original simple graph in the message passing layers by the generated virtual multigraph, that is,
\begin{align}
\begin{split}
         h_i^{l+1} = &\text{UPDATE}^{l}(h_i^{l}, r(i, i),  \\ & 
 \ \ \ \ \text{AGGREGATE}^l(h_j^{l}, r(i, j) \mid j \in \cN_i)).
\end{split}
\end{align}
For example, for GCN, APPNP and GCNII, we replace $\tilde{P}$ by $\hat{P},$ where 
$\hat{P}$ is defined as
\begin{align}
\label{eq:virtual_P}
\begin{split}
    \hat{P} = &\hat{D}^{-\frac{1}{2}}\hat{A}\hat{D}^{-\frac{1}{2}}, ~ \hat{A} = (\hat{A}_{ij}=z_{ij})%_{1\leq i, j \leq |V|}
    , \\ &\hat{D} = \text{diag}(\hat{D}_i=\sum_j z_{ij})%_{1\leq i \leq |V|}
    .
\end{split}
\end{align}
In addition, as the virtual multigraph already contains self-loops, there is no need to add the self-loops again to the message passing layers. This is different from conventional GNNs, where the self-loops are often added and forced to be a single edge. Though GIN \cite[]{xu2018} and JKNet \cite[]{xu2018a} also assign different weights for self-loops empirically, we are the first to propose a method to systematically determine the relative weights for self-loops and other edges.

In summary, conditional on the updated  parameters of DMPGM in each iteration, we sample the multiplicity of each edge. Then a set of multiedges and self-loops are generated from DMPGM, which can be used to build a virtual graph and update GNN trainable parameters.
We present the proposed EEGNN algorithm in Algorithm~\ref{alg:alg_1}.

\begin{algorithm}
\caption{EEGNN Algorithm}
\label{alg:alg_1}
\begin{algorithmic}
\STATE{Iterate Step~1 to Step~6 in Section~\ref{subsec:inference}} till the MCMC chains converge.\\
Set up initialization of trainable parameters in EEGNN.
\REPEAT 
    \STATE{
    1. Build the virtual graph and sample $\hat{P}$ according to (\ref{eq:virtual_P}),}\\
    \STATE{
    2. Use $\hat{P}$ to replace $\tilde{P}$,}\\
    \STATE{
    3. Update GNN parameters using the gradient descent,}\\
    \STATE{
    4. Obtain a new sampling for the parameters in DMPGM by implementing Step~1 to Step~6,}
\UNTIL {the convergence of the loss function of EEGNN}
\end{algorithmic}
\end{algorithm}

It is worth noting that we opted not to employ the stochastic block model in our study as it produces only dense graphs where the number of edges increases proportionally to the square of the number of nodes, whereas real-world networks tend to be sparse \cite[]{caron2017}. Moreover, the stochastic block model does not allow for constructing a virtual multi-graph on edges. To build the virtual multi-graph, it is needed to use a statistical model on edges instead of on nodes. 

\subsection{Comparison with Other Methods}
Our proposed method, EEGNN, addresses the performance deterioration of deep GNNs by using the structural information extracted from a Bayesian nonparametric graph model, DMPGM, to improve the performance of various deep message-passing GNNs. This is in contrast to relevant methods such as the attention and edge-label guided GNNs \cite[]{zhou2022attention} and edge-enhanced graph convolution networks \cite[]{cui2020}, which focus on integrating syntactic dependency or dependency label information into GCN to perform event detection or named entity recognition, respectively. Edge-feature-enhanced GNNs \cite[]{gong2019exploiting}, another competing method, focuses on integrating edge features instead of extracting edge information based on the observed graph in an unsupervised-learning fashion. Our EEGNN framework differs from these methods as it addresses the issue of \textit{mis-simplification} in deep GNNs and uses structural information from DMPGM to improve performance.

\section{EXPERIMENTS}
\subsection{Datasets and Bayesian Estimation}
In this section, we demonstrate through real data examples that EEGNN can effectively use the edge structural information to improve the performance for various GNNs. We conduct empirical experiments to compare EEGNN with representative baselines across six well-established network datasets. First, \textit{Cora}, \textit{Citeseer}, and  \textit{PubMed} are standard benchmark datasets for citation networks \cite[]{yang2016}. In these networks, nodes represent papers, and edges indicate cross citations between papers. Node features are the bag-of-words embedding of the contents, and node labels are academic subjects. 
Second, \textit{Texas}, \textit{Cornell}, and \textit{Wisconsin} are webpage cross-link networks \cite[]{pei2020}. Their nodes represent web pages of universities, and edges represent hyperlinks between them. Node features are bag-of-words embedding of the websites. Node labels contain five categories for the webs including students, projects, courses, staff, and faculty. Statistics for these datasets are summarized in  in Table~\ref{tab:data_statistics}.

\begin{table}[ht!]
\setlength\tabcolsep{1.5pt}
\caption{Graph datasets statistics. %Relative improvements of CATVI in percentage over OVI for HDP model are shown in parentheses.
}
\label{tab:data_statistics}
%\begin{subtable}
%\begin{center}
%\begin{small}
%\begin{sc}
\begin{tabularx}{0.485\textwidth}{lcccccc}
\toprule
Dataset &  \textit{Cora}&  \textit{Citeseer}  & \textit{PubMed} & \textit{Texas} & \textit{Wisconsin} & \textit{Cornell} \\ %\hline
\midrule
Nodes & 2,708 & 3,327 & 19,717 & 183 & 183 & 183\\
Edges & 5,429 & 4,732 & 44,338 & 309 & 499 & 295 \\
Degrees & 3.88 & 2.84 & 4.50 & 3.38 & 5.45 & 3.22 \\ 
Features &  1,433 & 3,703& 500& 1,703 & 1,703 & 1,703\\
Classes &  7 & 6 & 3 & 5 & 5& 5\\
\bottomrule
\end{tabularx}
%\end{sc}
%\end{small}
%\end{center}
%\end{subtable}
%\vspace{-0.1in}
\end{table}

\begin{figure}[t!]
\centering
\begin{subfigure}{0.43\textwidth}
\centering
\includegraphics[width=\textwidth]{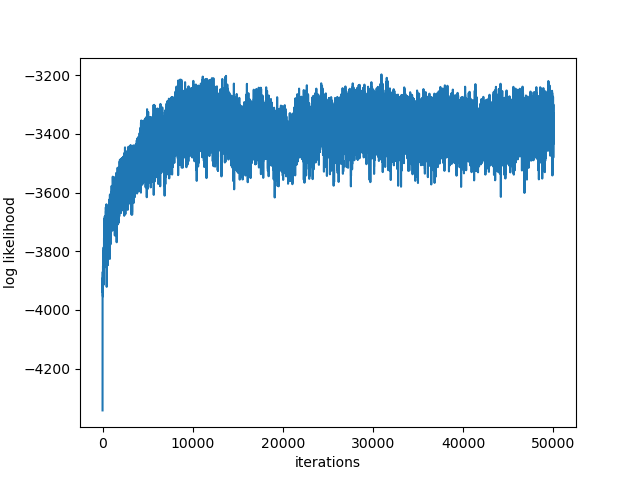}
\caption{The training log likelihood of DMPGM}
\label{fig:fig_texas_a}
\end{subfigure}
\centering
\begin{subfigure}{0.43\textwidth}
\centering
\includegraphics[width=\textwidth]{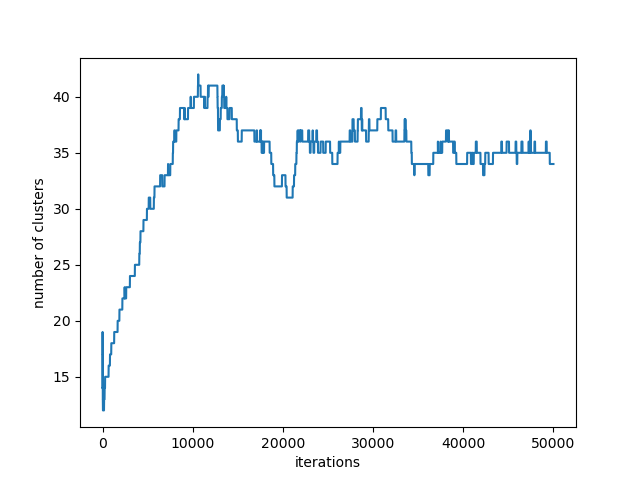}
\caption{The inferred number of edge clusters}
\label{fig:fig_texas_b}
\end{subfigure}
\centering
\begin{subfigure}{0.43\textwidth}
\centering
\includegraphics[width=\textwidth]{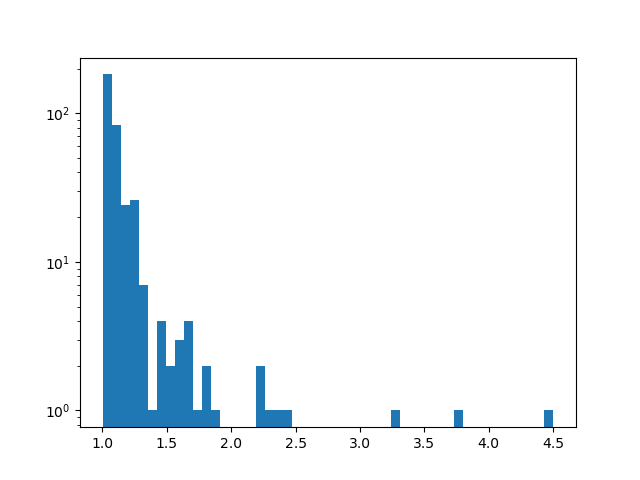}
\caption{Histogram of the expected multiplicity of virtual edges}
\label{fig:fig_texas_c}
\end{subfigure}
\caption{The MCMC inference results for \textit{Texas}. }
\label{fig:texas}
\end{figure}

Our experiments are implemented by using a gamma process as the completely random measure in (\ref{eq:construction}).
Following Section~\ref{subsec:inference}, we infer the parameters of DMPGM using MCMC in the following way. We use population based training \cite[]{jaderberg2017} to tune the hyperparameters in DMPGM. For each dataset, we grow the MCMC chain up to 50,000 epochs. 
Figures~\ref{fig:fig_texas_a} and \ref{fig:fig_texas_b} display the log-likelihood and number of clusters with respect to training epochs for the \textit{Texas} dataset. 
(The training results for the other datasets are shown in Appendix~\ref{app:mcmc_training}.)
Figure~\ref{fig:fig_texas_a} shows that the log-likelihood of \textit{Texas} for the DMPGM converges after 10,000 epochs. Moreover, benefiting from the Bayesian nonparametric model, we can infer the number of edge clusters in a data-adaptive manner \cite[]{liu2020}. 
Figure~\ref{fig:fig_texas_b} shows that the inferred number of edges per node (termed as \textit{multiplicity} of virtual edges per node) rises from an initial value of 10 to 50 at the start of training and then converges to around 35. 
The inferred edge multiplicity is displayed in the histograms in Figure~\ref{fig:fig_texas_c}. These histograms show that a large proportion of the edges in the observed graph have underlying multi-edges, suggesting the \textit{mis-simplification} in the original observed graph.
%We note that the clustering for edges in DMPGM is not the concept of the clustering for nodes in GNN. 

%  using each dataset

\subsection{Comparison with Baselines}
With the inference results of DMPGM, following Section~\ref{subsec:eemm}, we implement the experiments to compare  baseline GNNs and their edge enhanced versions. For the baseline GNNs, we chose SGC \cite[]{wu2019} and its variant, including APPNP \cite[]{Klicpera2019} and GCNII \cite[]{chen2020}, and hence name their edge enhanced versions as EE-SGC, EE-APPNP and EE-GCNII, respectively. To make a fair comparison, we follow the settings of the `sweet point' GNN hyperparameter configuration in \cite{chen2021} for all datasets. The details of these hyperparameter settings are collected in Appendix~\ref{app:hp}. For all experiments, the GNNS are trained with a maximum of 1000 epochs and an early stopping patience of 100 epochs. To analyze the effect of EEGNN with different numbers of layers, we run the experiments for 2, 16, 32 and 64 layers. 
We randomly split node features in each dataset into training and test sets, train the baseline GNNs and the edge enhanced versions using the same training set, and then compute the node clustering accuracy on the test set. In the transductive learning framework for GNNs, it is noted that the edge information is not partitioned as described in \cite{kipf2017a}. We repeat this procedure 50 times for each model and dataset. The mean predictive accuracy and the corresponding standard deviation are reported in Table~\ref{tab:all_results}.
%In Table~\ref{tab:all_results}, we present the mean predictive accuracy and the corresponding standard deviation over different train-test data split from 50 independent repetitions.

%To show that EEGNN can improve the performance of deep GNNs, we set all trained GNNs to have 32 layers.
%The average performance of each experiment is summarized in Table~\ref{tab:comparision_result}, along with the standard deviations of the node classification accuracy.

We observe a few apparent patterns. First, EEGNN can improve the performance of the baseline models in most cases. For example, SGC, the backbone GNN for various models, performs poorly with 32 layers (see Table~\ref{tab:32_layers}). However, with the aid of our EEGNN framework, the accuracy of the SGC model is increased by more than $6\%$ for \textit{Cora}, and by approximately $2\%$ across other candidate datasets. Moreover, SGC performs even worse with 64 layers for \textit{Cora} and \textit{Pubmed} (see Table~\ref{tab:64_layers}). EEGNNs largely improve the prediction accuracy in both cases,  by $9.89\%$ and $23.30\%$, respectively. It is worth noting that the improvements are attained without changing any other settings. As using virtual multigraph or observed simple graph brings in the only difference, this provides strong evidence to reveal that EEGNN can be used as a tool to enhance baseline GNNs by alleviating the \textit{mis-simplification} problem.

Second, for APPNP and GCNII, EEGNNs achieve similar accuracies on the \textit{Cora}, \textit{Citeseer} and \textit{PubMed} datasets, but substantially improve the performance on \textit{Texas}, \textit{Wisconsin} and \textit{Cornell}. Especially, with 64 layers, EE-GCNII for \textit{Texas} leads to more than $6\%$ improvement, and EE-APPNP for \textit{Citeseer} results in more than $10\%$ increase in the predictive accuracy. On the other hand, as APPNP and GCNII have already reached relatively high accuracy (approximately $70\%-80\%$) on the \textit{Cora}, \textit{Citeseer} and \textit{PubMed}, further enhancement to higher accuracy tends to be difficult. 

Finally, we observe that EEGNN has a larger impact on the performance of deeper SGC on the \textit{Cora}, \textit{Citeseer} and \textit{PubMed}. With only 2 layers, edge enhanced versions behave slightly worse than baseline models. However, with 32 or 64 layers, EEGNNs achieve considerable improvements. This is because the \textit{mis-simplification} applies to all layers. Therefore, the distortion of edge structural information is accumulated from the first to the last layer, resulting in severe performance deterioration.

\begin{table*}[ht!]
\centering
\caption{Results on real datasets: mean accuracy (\%) $\pm$ standard deviation (\%)
}
\label{tab:all_results}

\begin{subtable}{0.88\linewidth}
\caption{Number of layers: 2}
\centering
\label{tab:2_layers}

\begin{tabular}{lcccccc}
\toprule
&  \textit{Cora}&  \textit{Citeseer}  & \textit{PubMed} & \textit{Texas} & \textit{Wisconsin} & \textit{Cornell} \\
\midrule
SGC & \textbf{77.01$\pm$0.34}   &  69.18$\pm$0.35  & 75.46$\pm$0.28    &  56.16$\pm$4.99 &  48.59$\pm$6.59  &  57.84$\pm$2.76 \\
EE-SGC & 76.78$\pm$0.29   &  \textbf{69.60$\pm$0.37}  & \textbf{75.80$\pm$0.21}   &  \textbf{61.24$\pm$6.48} &  \textbf{53.45$\pm$8.95}  &  \textbf{58.92$\pm$3.55} \\
\addlinespace[0.8ex]
APPNP &  \textbf{82.22$\pm$0.39}   &  \textbf{71.73$\pm$0.76}  & \textbf{79.41$\pm$0.48}    &  61.41$\pm$5.27 &  52.55$\pm$7.44  &  57.73$\pm$2.74 \\
EE-APPNP & 81.48$\pm$0.47   &   71.45$\pm$0.54 & 
78.90$\pm$0.52    &  \textbf{66.80$\pm$3.74} &  \textbf{66.23$\pm$2.93}  &  \textbf{60.17$\pm$6.00} \\
\addlinespace[0.8ex]
GCNII &\textbf{82.21$\pm$0.67}   &  67.65$\pm$0.96  & \textbf{77.91$\pm$1.71}    &61.35$\pm$8.18 & 
\textbf{72.51$\pm$4.91} &  74.22$\pm$8.75 \\
EE-GCNII & 81.94$\pm$0.51   &  \textbf{81.48$\pm$0.59}  & 77.30$\pm$0.97    &  \textbf{64.22$\pm$9.02} &  70.94$\pm$6.10  &  \textbf{75.68$\pm$9.78} \\
\bottomrule
\end{tabular}

\end{subtable}

\vspace{0.5em}
\begin{subtable}{0.88\linewidth}
\caption{Number of layers: 16}
\centering
\label{tab:16_layers}
\begin{tabular}{lcccccc}
\toprule
&  \textit{Cora}&  \textit{Citeseer}  & \textit{PubMed} & \textit{Texas} & \textit{Wisconsin} & \textit{Cornell} \\
\midrule
SGC & \textbf{73.11$\pm$0.43}   &  67.79$\pm$0.56  & 70.45$\pm$0.17    &  56.27$\pm$4.92 &  48.63$\pm$6.62  &  57.84$\pm$2.76 \\
EE-SGC & 73.07$\pm$0.34   &  \textbf{68.55$\pm$0.35}  & \textbf{70.60$\pm$0.25}    &  \textbf{59.96$\pm$6.46} &  \textbf{50.59$\pm$8.72}  &  \textbf{57.84$\pm$2.76} \\
\addlinespace[0.8ex]
APPNP & \textbf{83.70$\pm$0.20}  &  72.51$\pm$0.52  & \textbf{80.42$\pm$0.30}    &  60.76$\pm$5.05 &  53.29$\pm$7.09  &  57.68$\pm$2.79 \\
EE-APPNP & 83.47$\pm$0.66   &  \textbf{73.20$\pm$0.92}  & 77.90$\pm$0.36    &  \textbf{66.08$\pm$4.62} &  \textbf{66.08$\pm$3.17}  &  \textbf{61.30$\pm$7.18} \\
\addlinespace[0.8ex]
GCNII & \textbf{84.77$\pm$0.37}   &  72.30$\pm$0.80  & 78.60$\pm$0.52    &  66.38$\pm$8.69 &  70.71$\pm$2.40  &  74.49$\pm$8.98 \\
EE-GCNII & 84.10$\pm$0.57   &  \textbf{72.50$\pm$1.40} & \textbf{78.81$\pm$0.66}    &  \textbf{73.30$\pm$3.85} &  \textbf{78.94$\pm$4.90}  &  \textbf{75.24$\pm$8.08} \\
\bottomrule
\end{tabular}

\end{subtable}

\vspace{0.5em}
\begin{subtable}{0.88\linewidth}
\caption{Number of layers: 32}
\centering
\label{tab:32_layers}
\begin{tabular}{lcccccc}
\toprule
&  \textit{Cora}&  \textit{Citeseer}  & \textit{PubMed} & \textit{Texas} & \textit{Wisconsin} & \textit{Cornell} \\
\midrule
SGC & 59.94$\pm$0.56   &  66.17$\pm$0.50  & 68.97$\pm$0.19    &  56.16$\pm$4.99 &  48.59$\pm$6.59  &  57.84$\pm$2.76 \\
EE-SGC & \textbf{ 66.46$\pm$0.83 }  &   \textbf{67.68$\pm$0.44}  & \textbf{70.68$\pm$0.68}    & \textbf{ 59.46$\pm$6.16 }&  \textbf{ 50.59$\pm$8.72 } & \textbf{ 58.92$\pm$3.15 } \\
\addlinespace[0.8ex]
APPNP &  83.55$\pm$0.50   &   72.11$\pm$0.64  & 80.22$\pm$0.34    &  61.57$\pm$5.28 &   52.71$\pm$7.34  &  57.78$\pm$2.75  \\
EE-APPNP & \textbf{ 83.79$\pm$0.39 }  &   \textbf{72.47$\pm$0.53}  & 79.23$\pm$0.27    & \textbf{66.00$\pm$4.33}&  \textbf{ 66.39$\pm$3.09 } & \textbf{61.84$\pm$7.56 } \\
\addlinespace[0.8ex]
GCNII &  85.34$\pm$0.53   & 73.26$\pm$0.86    & \textbf{ 79.89$\pm$0.33 }  &   70.49$\pm$5.48  & 69.06$\pm$2.70 &  74.05$\pm$8.56 \\
EE-GCNII & \textbf{ 85.70$\pm$0.41 }  &   \textbf{73.45$\pm$1.40}  & 79.72$\pm$0.43  & \textbf{ 75.24$\pm$3.72 }&  \textbf{ 79.37$\pm$0.43 } & \textbf{ 74.86$\pm$7.84 } \\
\bottomrule
\end{tabular}

\end{subtable}

\vspace{0.5em}
\begin{subtable}{0.88\linewidth}
\caption{Number of layers: 64}
\centering
\label{tab:64_layers}
\begin{tabular}{lcccccc}
\toprule
&  \textit{Cora}&  \textit{Citeseer}  & \textit{PubMed} & \textit{Texas} & \textit{Wisconsin} & \textit{Cornell} \\
\midrule
SGC & 25.65$\pm$1.93   &  63.08$\pm$0.52  & 40.98$\pm$1.73    &  56.16$\pm$4.99 &  48.55$\pm$6.58  &  57.84$\pm$2.76 \\
EE-SGC & \textbf{ 35.54$\pm$1.36 }  &   \textbf{65.42$\pm$0.17}  & \textbf{64.28$\pm$0.82}    & \textbf{ 59.46$\pm$6.16}&  \textbf{ 50.31$\pm$8.42 } & \textbf{ 58.92$\pm$3.15 } \\
\addlinespace[0.8ex]
APPNP &  83.58$\pm$0.49   &   72.10$\pm$0.48  & \textbf{80.42$\pm$0.42}    & 61.19$\pm$5.29 &  53.06$\pm$7.10  &  57.68$\pm$2.74  \\
EE-APPNP & \textbf{ 83.76$\pm$0.41 }  &   \textbf{72.16$\pm$0.65}  & 79.94$\pm$0.22    & \textbf{ 66.00$\pm$4.08 }&  \textbf{ 66.63$\pm$3.12 } & \textbf{ 61.75$\pm$7.43 } \\
\addlinespace[0.8ex]
GCNII & 85.46$\pm$0.31   &   \textbf{73.44$\pm$1.00}  & \textbf{80.08$\pm$0.37}    &  69.57$\pm$5.70 &   68.63$\pm$1.05  &  73.19$\pm$8.83  \\
EE-GCNII & \textbf{85.54$\pm$0.59}   &   72.24$\pm$1.26  & 79.93$\pm$0.46      & \textbf{ 75.62$\pm$3.65 }&  \textbf{ 76.57$\pm$3.89 } & \textbf{ 73.26$\pm$7.39 } \\
\bottomrule
\end{tabular}

\end{subtable}
\vspace{-0.1in}
\end{table*}
%\textcolor{red}{you can use outperform, be superior to, be inferior to, improved accuracy, the improvement is prominent, perform better....}

\subsection{Application in Financial Data}
\begin{figure}[h]
\centering
\includegraphics[width=0.44\textwidth]{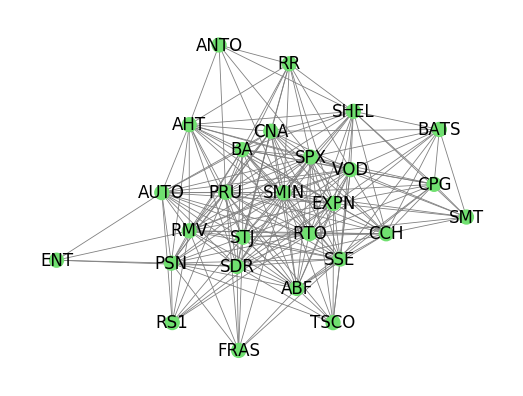}
\caption{The graph between FTSE UK 50 component stocks. Nodes in green denote individual stocks with their abbreviations in capital letters.}
\label{fig:stock_graph}
\end{figure}

\begin{figure}[t!]
\centering
\includegraphics[width=0.40\textwidth]{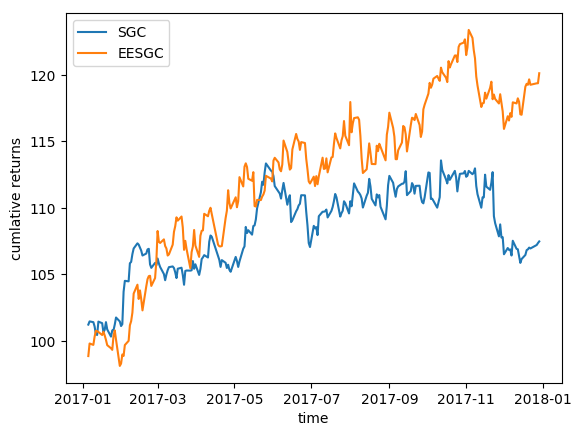}
\caption{The comparison of cumulative returns using SGC and EE-SGC.}
\label{fig:returns}
\end{figure}

GNN is widely used in the financial industry for the prediction of stock and bond prices \cite[]{wang2021, sharma2020, feng2022}. To evaluate the efficacy of  EEGNN in real-world financial data, we conduct an empirical study using EE-SGC to replace SGC in the current literature  and then make a comparison. We use the component stocks from the  `FTSE UK 50 index' with high capitalization and complete records between 2016-01-01 and 2017-12-31. We first construct the graph based on the Pearson correlations between stock returns, by connecting two stocks if their correlation is larger than 0.3. As shown in Figure~\ref{fig:stock_graph}, stocks, indicated by nodes are connected according to their pairwise correlation. Then, we build a learning pipeline using a sequential model of a long short-term memory (LSTM) network, SGC/EE-SGC, and a fully-connected layer. 
The model was trained using the data in 2016 and tested on the data in 2017. Moreover, the historical returns were used as input data, and the mean squared error between the model outputs and the realized next-day returns was used as the loss function.
The Long 20$\%$ strategy is adopted to build the portfolio as described in  \cite{pacreauGraphNeuralNetworks2021a}. For each trading day, we build a long only portfolio consisting of the top 20\% stocks
according to the predictive returns. The accumulated returns of the portfolio are shown in Figure~\ref{fig:returns}, where the initial portfolio value is set to be \$100. The results show that the portfolio constructed using EEGNN, which achieved better predictive accuracy, had higher returns.

\section{CONCLUSION}
\label{sec:conclusion}

This paper presents a novel explanation for the performance deterioration of deeper GNNs: \textit{mis-simplification}. We propose DMPGM, a Bayesian nonparametric graph model and its MCMC inference framework. Using the information obtained from DMPGM, we replace the original simple graph by the virtual graph, and use the virtual graph to aggregate the information in the graph. The experiments over various real datasets demonstrate that EEGNN can improve the performance of baseline GNN methods. Our paper paves a new way to use information extracted by statistical graph modelling to improve the performance of GNNs. One limitation of our proposal is that EEGNN only adds the virtual edges to the observed graph without removing edges according to the structural information. It is left for future work to develop a framework that allows to add and remove edges with the structural information simultaneously.

\subsubsection*{Acknowledgments}

We thank the anonymous reviewers for their useful comments during the review process.

%This work was performed during the first author's study at the London School of Economics.

Opinions expressed in this paper are those of the authors, and do not necessarily reflect the view of J.P.
Morgan. Opinions and estimates constitute our judgement as of the date of this Material, are for informational purposes only and are subject to change without notice. This Material is not the product of J.P.
Morgan’s Research Department and therefore, has not
been prepared in accordance with legal requirements
to promote the independence of research, including but
not limited to, the prohibition on the dealing ahead of
the dissemination of investment research. This Material is not intended as research, a recommendation,
advice, offer or solicitation for the purchase or sale of
any financial product or service, or to be used in any
way for evaluating the merits of participating in any
transaction. It is not a research report and is not intended as such. Past performance is not indicative
of future results. Please consult your own advisors
regarding legal, tax, accounting or any other aspects
including suitability implications for your particular
circumstances. J.P. Morgan disclaims any responsibility or liability whatsoever for the quality, accuracy or
completeness of the information herein, and for any reliance on, or use of this material in any way. Important
disclosures at: www.jpmorgan.com/disclosures.

\bibliographystyle{apa}
\bibliography{references}

\clearpage

\onecolumn 
\thispagestyle{empty}
\hsize\textwidth
    \linewidth\hsize \toptitlebar {\centering
        {\Large\bfseries Supplementary Material for `EEGNN: Edge Enhanced Graph Neural Network with a Bayesian Nonparametric Graph Model'  \par}}
    \bottomtitlebar

\appendix

This supplementary material contains a short review of completely random measures in Appendix~\ref{app:cmr}, the details of MCMC steps in Appendix~\ref{app:mcmc_details}, technical proofs and
derivations in Appendix~\ref{app:sec_proof}, the results of MCMC for several datasets in Appendix~\ref{app:mcmc_training}, hyperparameter settings in Appendix~\ref{app:hp}, and computational complexity analysis and code in Appendix~\ref{app:codes}.

\section{Completely Random Measure}
\label{app:cmr}
% such that (i) $G \mapsto G(A)$ is $\cM$-measurable for any $A \in \cF$ and (ii) $A \mapsto G(A)$ is a measure for any realization of $G$ \citep{ghosal2017}. 
%Moreover, a random measure is called a completely random measure \citep{Kingman1993} if it also satisfies the condition that (iii) $P(A_i)$ is independent of $P(A_j)$ for any disjoint subsets $A_i$ and $A_j$ in $\Omega$. 
Completely random measures \citep{ghosal2017}, including gamma process, inverse Gaussian process and stable process, are commonly used as priors for infinite-dimensional latent variables in Bayesian nonparametric models, because their realizations are atomic measures with countable-dimensional supports.
Suppose that $(\Omega,\cF)$ is a Polish sample space, $\Theta$ is the set of all bounded measures on $(\Omega,\cF)$ and $\cM$ is a $\sigma$-algebra on $\Theta$. A complete random measure from $(\Theta, \cM)$ into $(\Omega,\cF)$
can be characterized by its Laplace transform \citep{Kingman1993},
\begin{equation*}
\E\big[e^{-tP(A)}\big] = \exp \Big\{- \int_{A} \int_{(0, \infty]} (1- e^{-t\pi})v^c (dx , ds) \Big\},
\end{equation*}
where $A$ is any measurable subset of $\Omega$ and $v^c(dx , ds)$ is called the intensity measure. If $v^c(dx , ds) = \kappa(dx) v(ds)$, where $\kappa(\cdot)$ and $v(\cdot)$ are measures on $\Omega$ and $(0, \infty]$, respectively, the completely random measure is homogeneous and  $v(\cdot)$ is called the L\'evy measure.
If $\int_{0}^{\infty} v(ds) = \infty$, the complete random measure is finite activity.

We can view this completely random measure as a Poisson process on the product space $\Omega \times (0, \infty]$ using the intensity measure and denote this completely random measure as $\text{CRM}(\kappa, v)$. For example, the gamma process $\gammaPro(\kappa)$ has L\'evy measure $v(ds) = s^{-1}e^{-s}ds$ such that $Q(A) \sim \Gamma(\kappa(A), 1)$ if $Q \sim \gammaPro(\kappa)$, where $\Gamma(\alpha, \beta)$ is a gamma distribution with density $\frac{\beta^\alpha}{\Gamma(\alpha)}x^{\alpha-1}e^{-\beta x}.$ Therefore, its normalization, Dirichlet process $P \sim \DirPro(\kappa)$ \citep{ferguson1973}
satisfies
\begin{equation*}
    \big(P(A_1),\dots ,P(A_n)\big) \sim \text{Dirichlet} \big( \kappa(A_1),\dots ,\kappa(A_n)\big)
\end{equation*}
for any partition $\varOmega = (A_1,\dots,A_n)$, where $\bigcup _{i=1}^{n}A_i = \Omega$ and $A_i \bigcap A_j = \emptyset$ for any $i$ and $j$.  Griffiths–Engen–McCloskey  (GEM)  distribution, which is the distribution of the weights
in a Dirichlet process. For $(\pi_1, \pi_2, \cdots) \sim \text{GEM}(\alpha)$, it can be sampled by $\pi_i = g_i \prod_{l=1}^{i-1} g_l$, where $g_i \sim \mathrm{B}\text{eta} (1, \alpha)$ independently  \citep{ghosal2017}.

\section{MCMC technical details and derivations}
\label{app:mcmc_details}
\setcounter{equation}{0}
\renewcommand\theequation{B.\arabic{equation}}

\subsection{Derivations for Step 1}
\label{app:step_1}

With the setup of DMPGM in~(\ref{eq:construction}) and the formula of moments for Dirichlet-multinomial distribution, we obtain that
\begin{equation*}
p(w_{0, 1}, \dots, w_{0, |V|} \mid  \widebar{w}_{0}, \bz, \bc) ~\propto~ \prod_{k=1}^{K}
\frac{\Gamma(\widebar{w}_0)}{\Gamma(\widebar{w}_0 + n_k)}
\prod_{i=0}^{N}\frac{\Gamma(w_{0, i} + n_{k,i})}{\Gamma(w_{0, i})} \cdot \prod_{i=1}^{N} v(w_{0, i}) \cdot u\big(\widebar{w}_0 -   \sum_{i=1}^{|V|}w_{0,i}\big),
\end{equation*}
where $n_{k,i}=\sum_{j=1}^{|V|} \sum_{l=1}^{z_{ij}} \{\mathds{1}_{c_{ijl}=k} +  \mathds{1}_{c_{jil}=k}\}$,  $v(\cdot)$ is the weight intensity measure for the complete random measure of $W_0$, and $u(\cdot)$ is the density function for $W_0(\Omega)$ that can be derived using its Laplace transform. 
To infer the posterior distributions for these parameters, we present the gradient of the log-posterior with respect to $w_0$, which will be used in Hamiltonian Monte Carlo,
\begin{equation*}
\label{eq:gradient_hmc}
\begin{gathered}
    \nabla_{w_{0,i}} \log p(w_{0,1}, \dots, w_{0, |V|} \mid \widebar{w}_{0}, \bz, \bc) =
    \sum_{i=1}^{|V|}
    \sum_{k=1}^{K} \big\{\Phi(n_{k, i} + w_{0, i}) - \Phi(w_{0, i})\big\} \\ +
    \sum_{i=1}^{|V|}
    \nabla_{w_{0,i}}\log v(w_{0, i})  + \nabla_{w_{0,i}}\log u(\widebar{w}_{0} - \sum_{i=1}^{|V|}w_{0,i}),
\end{gathered}
\end{equation*}
where $\Phi$ is the digamma function.

\subsection{Derivations for Step 6}
\label{add:step_6}
By the formulas of the densities for Poisson distribution and gamma distribution, we have that
\begin{equation*}
    p(\widebar{w_{k}} \mid \widebar{w}_{0}, \bpi, \bc, \bz) ~\propto~ \frac{(\pi_k \widebar{w}_{k}^2)^{n_k}e^{-\pi_k \widebar{w}_{k}^2}}{n_k!} \cdot
    \frac{1}{\Gamma(\widebar{w}_{0})} \widebar{w}_{k}^{\widebar{w}_{0}-1}
    e^{-\widebar{w}_{k}}.
\end{equation*}
Therefore, the log-posterior with respect to $\widebar{w}_{k}$ is
\begin{equation*}
\log p(\widebar{w}_{k} \mid \widebar{w}_{0}, \bpi, \bc, \bz)= ({2n_k+\widebar{w}_{0}-1}) \log \widebar{w}_{k} - \widebar{w}_{k} - \widebar{w}_{k}^2\pi_k + \text{constant}.
\end{equation*}
Similarly, following \cite{caron2017} and \cite{liu2020}, we obtain that
\begin{equation*}
    p(\widebar{w}_{0} \mid \widebar{w}_{k}, \bpi, \bc, \bz) ~\propto~ \prod_{k=1}^{K}
    \frac{1}{\Gamma(\widebar{w}_{0})} \widebar{w}_{k}^{\widebar{w}_{0}-1}
    e^{-\widebar{w}_{k}} \cdot u({\widebar{w}_{0}}). 
\end{equation*}
and hence the corresponding log-posterior is
\begin{equation*}
\log p(\widebar{w}_{0} \mid \widebar{w}_{k}, \bpi, \bc, \bz)= \log u({\widebar{w}_{0}}) +  {\widebar{w}_{0}} \sum_{k=1}^{K}\log ( \widebar{w}_{k})  - K\log\Gamma(\widebar{w}_{0}) + \text{constant}.
\end{equation*}

\section{Technical Proofs and Derivations}
\label{app:sec_proof}

\subsection{Proof of Theorem~{\ref{theo:theorem_1}}}
\label{add:proof_theorem_1}
The proof for Theorems 3 and 4 in \cite{caron2017} can be directly adapted to DMPGM, Therefore, we only provide a sketch of the proof. First, we show that Theorem 3 in \cite{caron2017} also holds for DMPGM. We use 
$$
\tilde{D}_{ij} \mid \{W_k\} \sim \text{Poisson}\big(\sum_{k} \pi_k W_k([i-1, i])W([j-1, j])\big),
$$
to replace (54) in Appendix C.2 of \cite{caron2017}. Consequently, (55) holds because for any $k$ we have
\begin{equation}
\label{app:eq_conv}
W_k([0, \alpha]) / W_0([0, \alpha]) = O(1) \ \  \text{almost surely as} \  \alpha \rightarrow \infty.
\end{equation}
Second, we show that Theorem 4 in \cite{caron2017} also holds for DMPGM. Specifically, (59) becomes 
$$
X_n \mid \{W_k^{(2)} \} \sim \text{Poisson}\Big[\frac{1}{2} \psi \big\{ W (\cS_n^{(2)})\big\}\Big], 
$$ so that (62) in \cite[]{caron2017} can be achieved by (\ref{app:eq_conv}).
Finally, we complete the proof of Theorem~\ref{theo:theorem_1} for DMPGM by keeping the remaining parts of the proof of Theorem 4 in \cite{caron2017} unchanged.

\section{Inference results for DMPGM}
\label{app:mcmc_training}
\begin{figure}[h]
\centering
\begin{subfigure}{0.32\textwidth}
\centering
\includegraphics[width=\textwidth]{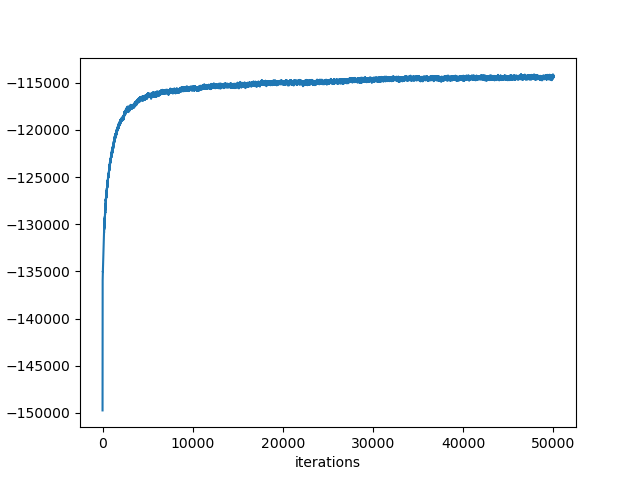}
\caption{\textit{Cora}}
\end{subfigure}
\begin{subfigure}{0.32\textwidth}
\centering
\includegraphics[width=\textwidth]{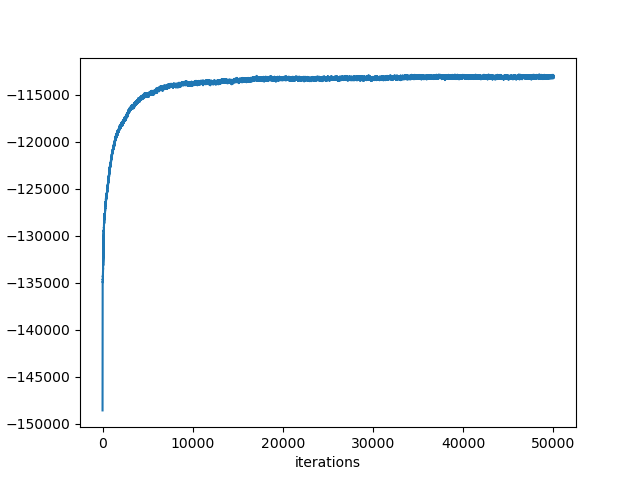}
\caption{\textit{Citeseer}}
\end{subfigure}
\begin{subfigure}{0.32\textwidth}
\centering
\includegraphics[width=\textwidth]{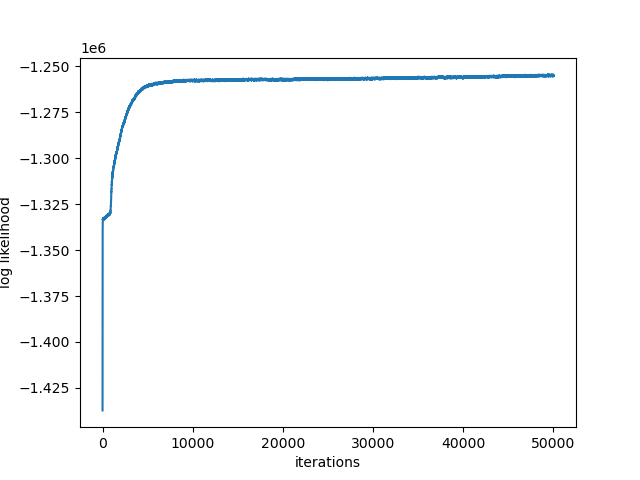}
\caption{\textit{PubMed}}
\end{subfigure}
\begin{subfigure}{0.32\textwidth}
\centering
\includegraphics[width=\textwidth]{pictures/TEXAS_log_likelihood}
\caption{\textit{Texas}}
\end{subfigure}
\begin{subfigure}{0.32\textwidth}
\centering
\includegraphics[width=\textwidth]{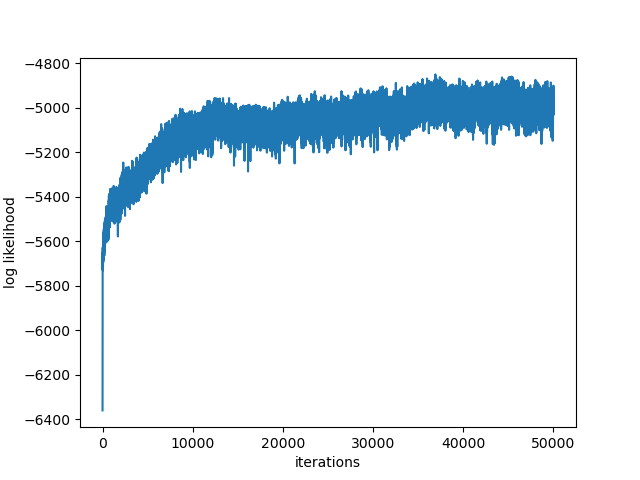}
\caption{\textit{Wisconsin}}
\end{subfigure}
\begin{subfigure}{0.32\textwidth}
\centering
\includegraphics[width=\textwidth]{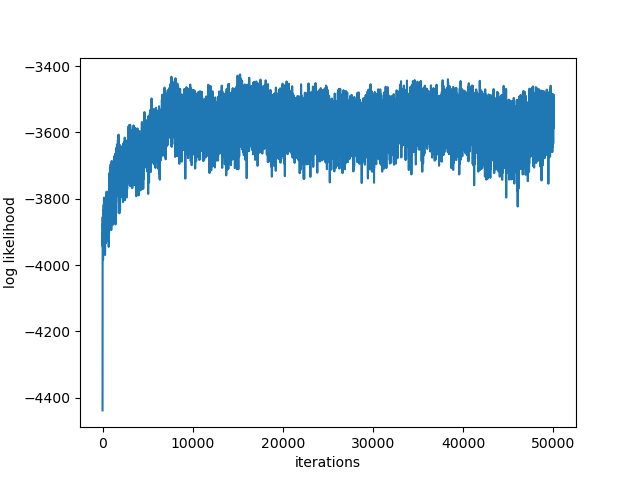}
\caption{\textit{Cornell}}
\end{subfigure}
\caption{Log-likelihood over the course of the MCMC chain for each dataset.}
\label{fig:log_likelihood}
\end{figure}

\begin{figure}
\centering
\begin{subfigure}{0.32\textwidth}
\centering
\includegraphics[width=\textwidth]{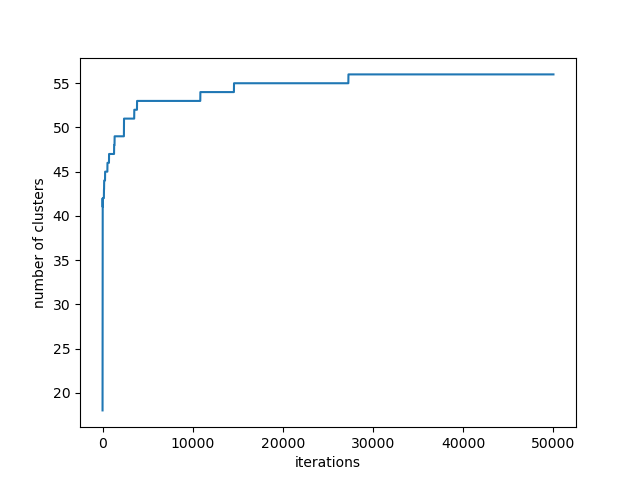}
\caption{\textit{Cora}}
\end{subfigure}
\begin{subfigure}{0.32\textwidth}
\centering
\includegraphics[width=\textwidth]{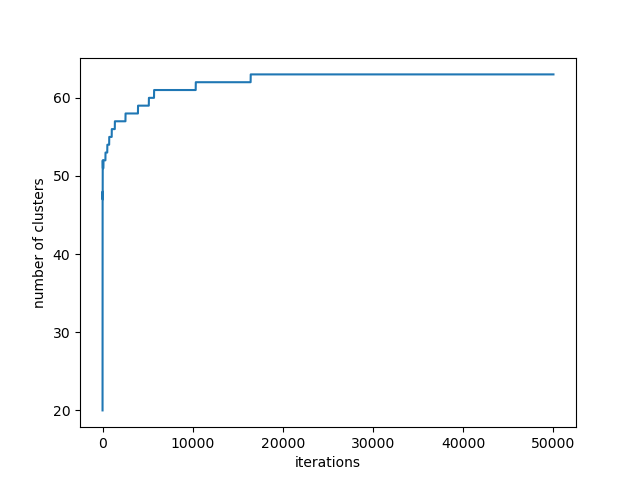}
\caption{\textit{Citeseer}}
\end{subfigure}
\begin{subfigure}{0.32\textwidth}
\centering
\includegraphics[width=\textwidth]{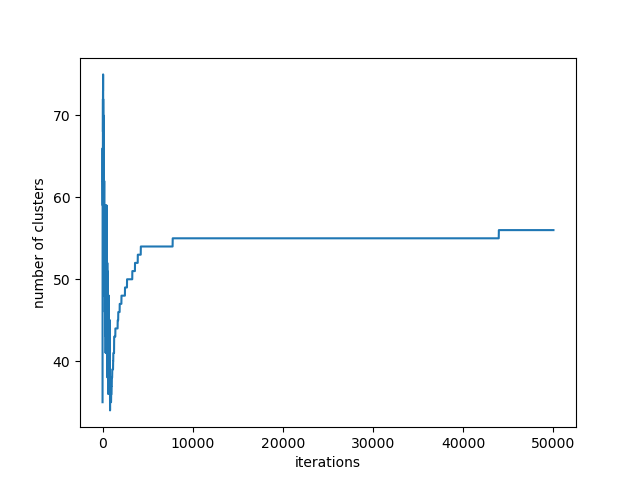}
\caption{\textit{PubMed}}
\end{subfigure}
\begin{subfigure}{0.32\textwidth}
\centering
\includegraphics[width=\textwidth]{pictures/TEXAS_active_k}
\caption{\textit{Texas}}
\end{subfigure}
\begin{subfigure}{0.32\textwidth}
\centering
\includegraphics[width=\textwidth]{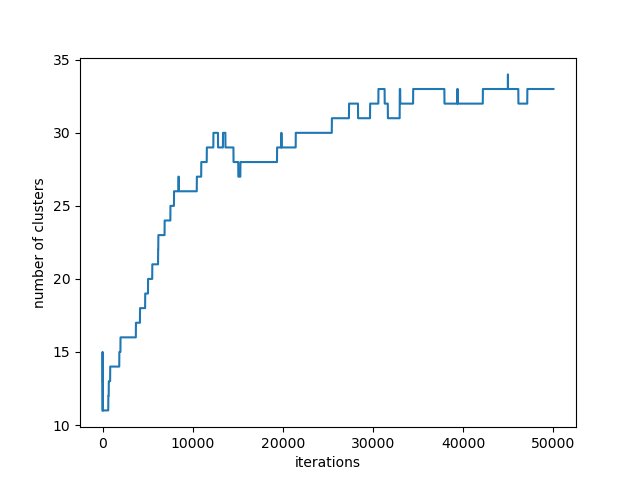}
\caption{\textit{Wisconsin}}
\end{subfigure}
\begin{subfigure}{0.32\textwidth}
\centering
\includegraphics[width=\textwidth]{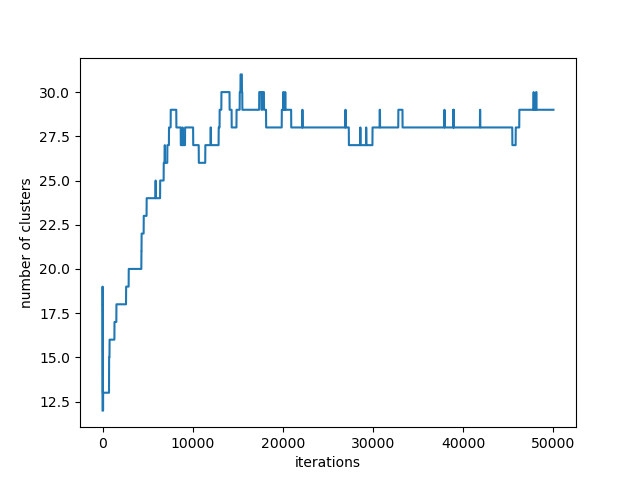}
\caption{\textit{Cornell}}
\end{subfigure}
\caption{Number of clusters inferred by DMPGM over the course of the MCMC chain.}
\label{fig:number_clusters}
\end{figure}

\begin{figure}
\centering
\begin{subfigure}{0.32\textwidth}
\centering
\includegraphics[width=\textwidth]{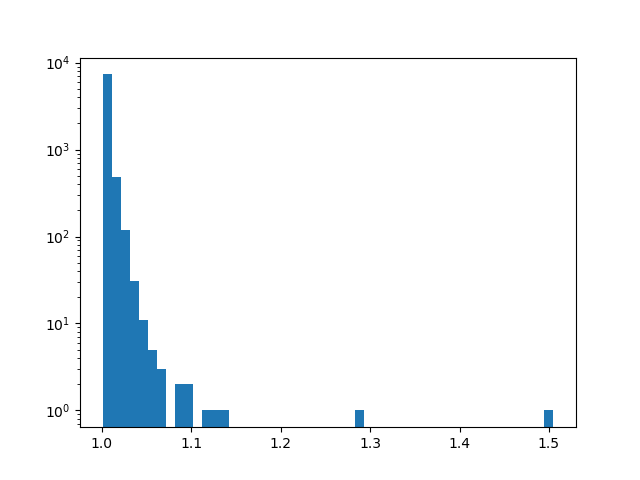}
\caption{\textit{Cora}}
\end{subfigure}
\begin{subfigure}{0.32\textwidth}
\centering
\includegraphics[width=\textwidth]{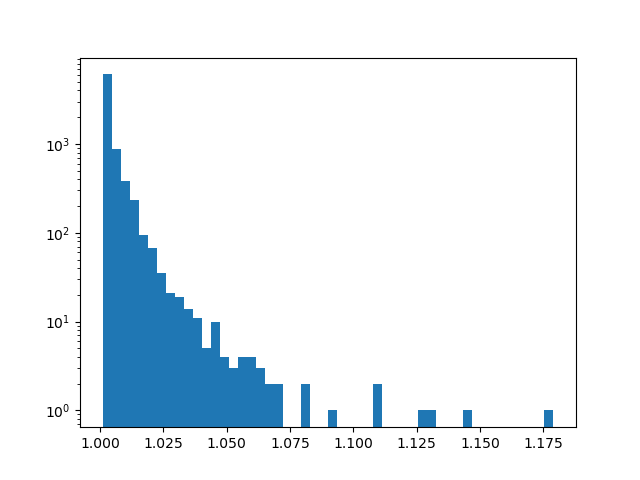}
\caption{\textit{Citeseer}}
\end{subfigure}
\begin{subfigure}{0.32\textwidth}
\centering
\includegraphics[width=\textwidth]{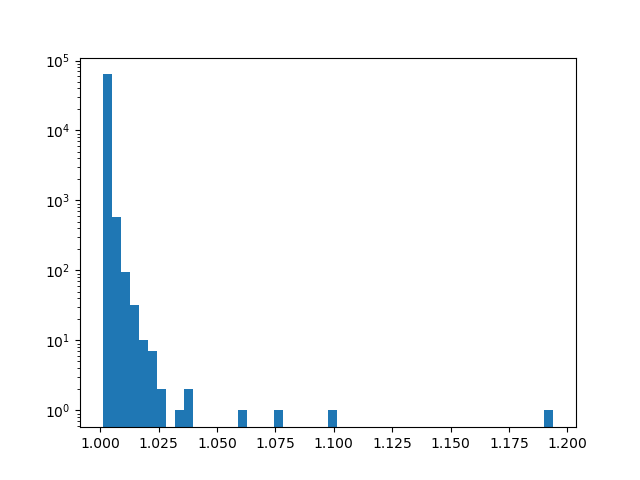}
\caption{\textit{PubMed}}
\end{subfigure}
\begin{subfigure}{0.32\textwidth}
\centering
\includegraphics[width=\textwidth]{pictures/TEXAS_hist}
\caption{\textit{Texas}}
\end{subfigure}
\begin{subfigure}{0.32\textwidth}
\centering
\includegraphics[width=\textwidth]{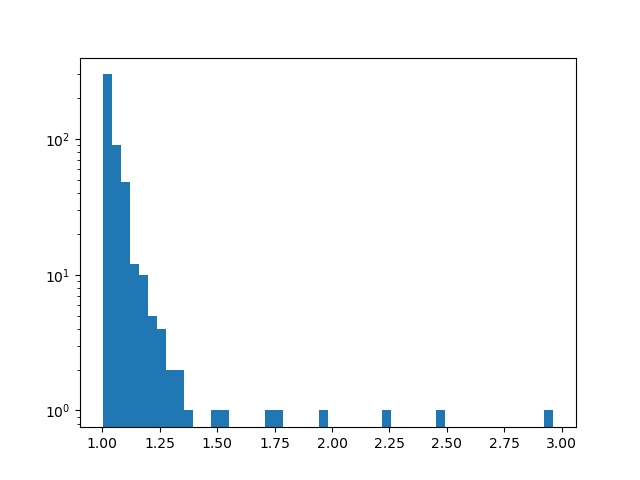}
\caption{\textit{Wisconsin}}
\end{subfigure}
\begin{subfigure}{0.32\textwidth}
\centering
\includegraphics[width=\textwidth]{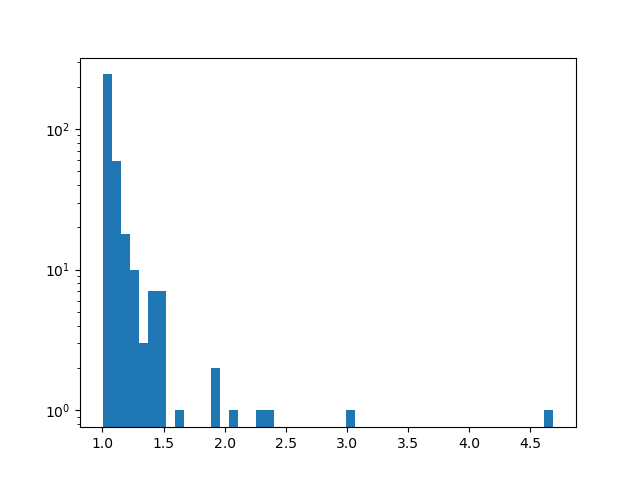}
\caption{\textit{Cornell}}
\end{subfigure}
\caption{Histograms of the expected multiplicity of virtual edges formed in the EEGNN framework using each dataset.}
\label{fig:multiplicity}
\end{figure}

The log likelihood and the number of edge clusters in the training process are shown in Figure~\ref{fig:log_likelihood} and Figure~\ref{fig:number_clusters}, respectively. The inferred edge multiplicity is shown in Figure~\ref{fig:multiplicity}.

\section{Hyperparameters}
\label{app:hp}
We list the hyperparameters used in our experiments in Table~\ref{tab:hp} below.

\begin{table}[h]
    \centering
    \begin{subtable}{1.0\linewidth}
    \centering
    
\setlength{\tabcolsep}{0.2em}

\begin{tabular}{lcccccc}
\toprule
&  \textit{Cora}&  \textit{Citeseer}  & \textit{PubMed} & \textit{Texas} & \textit{Wisconsin} & \textit{Cornell} \\

\midrule
dim\_hidden& 64 & 256 & 256 & 64 & 64 & 64 \\
alpha& 0.1 & 0.1 & 0.1 & 0.1 & 0.1 & 0.1 \\
weight\_decay& 0.0005 &  0.0005 & 0.0005 & 0.0001 & 0.0005 &  0.0005 \\
lr & 0.01 & 0.01 & 0.01 & 0.01  & 0.01 & 0.01 \\
dropout& 0.6 & 0.7 &  0.6 & 0.5 & 0.5 & 0.5 \\
\bottomrule
\end{tabular}

    \caption{Hyperparameters for SGC and EE-SGC.}
    \end{subtable}
    \begin{subtable}{1.0\linewidth}
    \centering
    
\setlength{\tabcolsep}{0.2em}

\begin{tabular}{lcccccc}
\toprule
&  \textit{Cora}&  \textit{Citeseer}  & \textit{PubMed} & \textit{Texas} & \textit{Wisconsin} & \textit{Cornell} \\

\midrule
  dim\_hidden & 64 & 64 & 64& 64& 64& 64\\
  alpha& 0.1 & 0.1& 0.1& 0.1& 0.1& 0.1\\
  lr & 0.01& 0.01& 0.01& 0.01& 0.01& 0.01\\
  dropout& 0.& 0.& 0.& 0.& 0.& 0.\\
  weight\_decay1& 0.005& 0.005& 0.005& 0.005& 0.005& 0.005\\
  weight\_decay2& 0.& 0.& 0.& 0.& 0.& 0.\\
  embedding\_dropout& 0.5& 0.5& 0.5& 0.5& 0.5& 0.5\\
\bottomrule
\end{tabular}

    \caption{Hyperparameters for APPNP and EE-APPNP.}
    \end{subtable}
    \begin{subtable}{1.0\linewidth}
    \centering
    
\setlength{\tabcolsep}{0.2em}

\begin{tabular}{lcccccc}
\toprule
&  \textit{Cora}&  \textit{Citeseer}  & \textit{PubMed} & \textit{Texas} & \textit{Wisconsin} & \textit{Cornell} \\

\midrule
  dim\_hidden& 64 & 256& 256 &64&64&64\\
  alpha& 0.1& 0.1& 0.1 &0.5&0.5&0.5\\
  lamda& 0.5& 0.6& 0.4 &1.5&1.0&1.0\\
  weight\_decay1& 0.01& 0.01&0.0005&0.0001&0.0005&0.0001\\
  weight\_decay2& 0.0005& 0.0005&0.0005&0.0001&0.0005&0.0001\\
  lr& 0.01& 0.01& 0.01& 0.01& 0.01& 0.01\\
  dropout& 0.6& 0.7& 0.6&0.5&0.5&0.5\\
\bottomrule
\end{tabular}

    \caption{Hyperparameters for GCNII and EE-GCNII.}
    \end{subtable}
    \caption{Hyperparameters in the training.}
    \label{tab:hp}
\end{table}

\section{Data and Code}
\label{app:codes}
%Python code for EEGNN is available at \url{https://anonymous.4open.science/r/eegnn-C3D7}. 
We obtained the datasets from the publically available source \url{https://pytorch-geometric.readthedocs.io/en/latest/modules/datasets.html}. All data do not contain personally identifiable
information or offensive content. We conducted our experiments on a c5d.4xlarge instance on the AWS EC2 platform,
with 16 vCPUs and 32 GB RAM. The codes for training conventional GNNs are from \url{https://github.com/VITA-Group/Deep\_GCN\_Benchmarking} under MIT license.

\clearpage
%\subsubsection*{Acknowledgments}
%Opinions expressed in this paper are those of the authors, and do not necessarily reflect the view of J.P. Morgan. Opinions and estimates constitute our judgement as of the date of this Material, are for informational purposes only and are subject to change without notice. This Material is not the product of J.P. Morgan’s Research Department and therefore, has not been prepared in accordance with legal requirements to promote the independence of research, including but not limited to, the prohibition on the dealing ahead of the dissemination of investment research. This Material is not intended as research, a recommendation, advice, offer or solicitation for the purchase or sale of any financial product or service, or to be used in any way for evaluating the merits of participating in any transaction. It is not a research report and is not intended as such. Past performance is not indicative of future results. Please consult your own advisors regarding legal, tax, accounting or any other aspects including suitability implications for your particular circumstances. J.P. Morgan disclaims any responsibility or liability whatsoever for the quality, accuracy or completeness of the information herein, and for any reliance on, or use of this material in any way. Important disclosures at: www.jpmorgan.com/disclosures.

\end{document}